\documentclass[runningheads]{llncs}
\usepackage[T1]{fontenc}
\usepackage{graphicx}
\usepackage{booktabs}
\usepackage{adjustbox}
\usepackage{color,soul}

\begin{document}
\title{Visual In-Context Learning for Few-Shot Eczema Segmentation}
%
%

\author{Neelesh Kumar, Oya Aran, Venugopal Vasudevan}
\authorrunning{Kumar et al.}

%
\institute{Procter and Gamble, Mason, OH 45040 USA \\
\email{\{kumar.n.40, aran.o, venu\}@pg.com}}
%
\maketitle              
\begin{abstract}
Automated diagnosis of eczema from digital camera images is crucial for developing applications that allow patients to self-monitor their recovery. An important component of this is the segmentation of eczema region from such images. Current methods for eczema segmentation rely on deep neural networks such as convolutional (CNN)-based U-Net or transformer-based Swin U-Net. While effective, these methods require high volume of annotated data, which can be difficult to obtain. Here, we investigate the capabilities of visual in-context learning that can perform few-shot eczema segmentation with just a handful of examples and without any need for retraining models. Specifically, we propose a strategy for applying in-context learning for eczema segmentation with a generalist vision model called SegGPT. When benchmarked on a dataset of annotated eczema images, we show that SegGPT with just 2 representative example images from the training dataset performs better (mIoU: 36.69) than a CNN U-Net trained on 428 images (mIoU: 32.60). We also discover that using more number of examples for SegGPT may in fact be harmful to its performance. Our result highlights the importance of visual in-context learning in developing faster and better solutions to skin imaging tasks. Our result also paves the way for developing inclusive solutions that can  cater to minorities in the demographics who are typically heavily under-represented in the training data. 
\end{abstract}
\section{Introduction}
Eczema is one of the most common skin disorders, with over 10\% of the population affected by it in the United States alone \cite{abuabara2019prevalence}. While there can be no substitute for expert dermatologists, automated diagnosis of the disease can enable patients to self-monitor their recovery using self-acquired images, potentially leading to more effective recovery due to the psychological effects \cite{horne1989preliminary,ehlers1995treatment}. An important part of this automated analysis is developing a sufficiently robust algorithm that can accurately segment the eczema region from the patient-acquired digital camera images. \cite{hurault2022detecting,ch2014segmentation,roy2019skin}. 

The state-of-the-art approaches for automated eczema segmentation rely on deep neural networks (DNN) for supervised learning, with the most popular architecture being a convolutional neural network-based U-Net (CNN U-Net) \cite{nisar2023eczema,anand2023fusion}. More recent approaches, such as Swin U-Net, leverage attention-based transformers that can capture long-range dependencies \cite{cao2022swin}. These methods have demonstrated remarkable performance improvements over traditional approaches that relied on hand-crafted visual features \cite{ch2014segmentation,schnurle2017using}. However, a common theme across all these methods is the need for sufficient data to train the DNN, which increases with the increasing complexity of the network architecture \cite{adadi2021survey}. 

For applications such as skin lesion segmentation that require expert annotations, acquiring a large dataset can be prohibitive from the point of view of both cost and time \cite{shi2019active}. This is further exacerbated by the need for a diverse dataset covering all sources of variations such as skin tone, images from different body parts, and varying levels of disease \cite{seth2017global,hurault2022detecting}. Current methods to train high-capacity networks using limited data employ transfer learning techniques such as domain adaptation, knowledge distillation, or finetuning \cite{tan2018survey,farahani2021brief}. While these techniques have proven to be effective for many learning tasks, including skin segmentation \cite{al2020comparative}, these methods still require enough labeled and unlabeled training data to work well. 

A long-standing goal in artificial intelligence is to learn task-agnostic representations that can be used across various tasks without requiring further learning \cite{goertzel2014artificial}. This few/zero-shot form of learning eliminates the necessity of training task-specific models and hence any need for training data. While the goal seems ambitious, we are already seeing an outpour of results from the Natural Language Processing (NLP) community \cite{radford2019language,liu2023summary}. Known popularly as large language models (LLMs), these models learn task-agnostic representations through pretraining on a large corpus of text data, and exhibit remarkable generalization on NLP downstream applications\cite{liu2023summary}. The operating principle is In-context Learning, where a few domain-specific input-output pairs are provided as in-context examples (prompts) to the model, along with the input test example \cite{zhang2023makes,radford2021learning}. The prompt and the input test example are used to predict the corresponding output for the test example without updating model weights \cite{zhang2023makes}. 

A growing body of work is now attempting to replicate this success of NLP for vision tasks using a paradigm known as visual in-context learning  \cite{bar2022visual,zhang2023makes,wang2023images,wang2023seggpt,oquab2023dinov2,kirillov2023segment}. The idea is similar to that in NLP: a high-capacity network such as a Vision Transformer (ViT) \cite{dosovitskiy2020image} is trained on a large corpus of images using Masked Image Modeling (MIM) \cite{xie2022simmim,he2022masked}. The complexity of the MIM task, i.e., predicting large missing patches in the image, forces the network to gain semantic and contextual understanding of the images resulting in it learning powerful general representations \cite{he2022masked}. These learned representations are versatile enough to perform various downstream vision tasks such as keypoint detection, segmentation, depth estimation, etc., following prompt-style input representation \cite{wang2023images,wang2023seggpt}. With visual in-context learning, it is possible to perform few-shot segmentation of eczema images without requiring an abundance of training data. 

\begin{figure*}[t!]
    \centering
    \includegraphics[width=\textwidth]{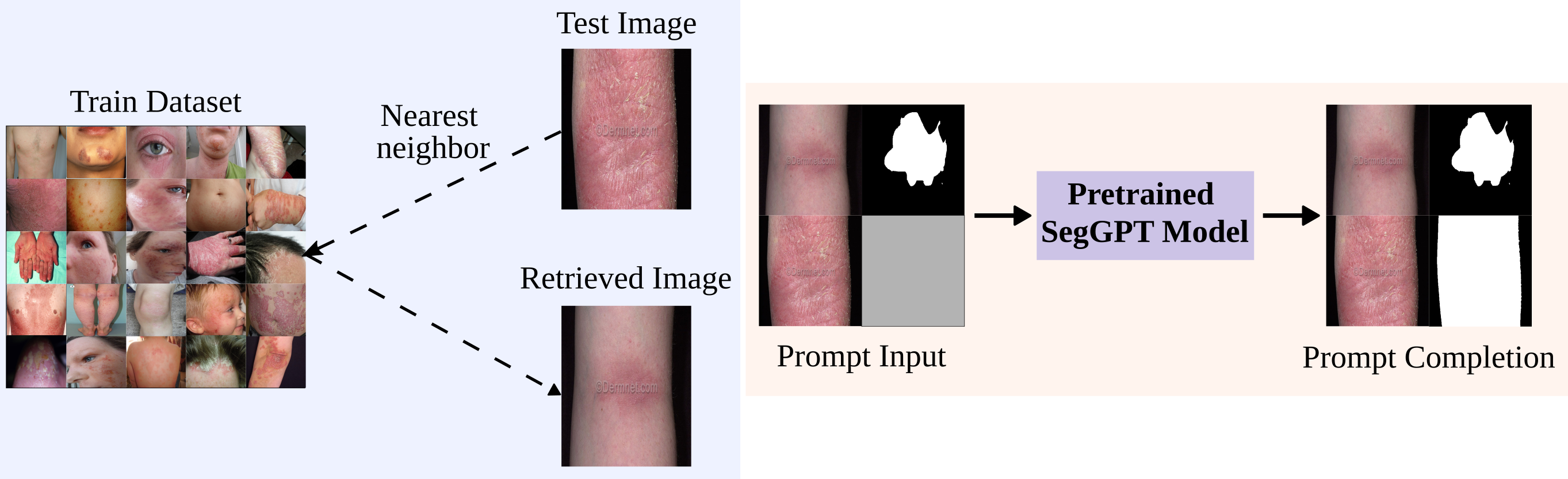}
    \caption{\textbf{Overview. }For each test image, its $k$-nearest neighbors are retrieved from the training dataset. The retrieved images and their masks, along with the test image are stitched together to construct a prompt that is fed to a pretrained SegGPT model. The task of the model is to predict the missing output image for the given test image.}
    \label{fig:fig1}
\end{figure*}

This work presents an automated approach to few-shot eczema segmentation using visual in-context learning (Figure \ref{fig:fig1}). Specifically, we employ a pretrained generalist vision model called SegGPT \cite{wang2023seggpt} and evaluate it for eczema segmentation on a dataset of skin images acquired from the web and a consumer study. We report that SegGPT with just two example images in the prompt performs better (mIoU: 36.69) than a CNN U-Net trained on 428 images (mIoU: 32.60). The key to this result is the strategy for prompt selection: the examples in the prompt must be representative of the task. Our simple framework for prompt selection retrieves nearest neighbors of the test image from the training dataset, and uses them as examples to construct the prompt. We also discover that the performance of SegGPT is strongly dependent on the number of examples in the prompt, but surprisingly does not have a linear relation. Rather, the performance increases up to a certain number $k<4$, and then starts decreasing. Our results highlight the promise of foundational generalist vision models in developing faster and better solutions for skin lesion segmentation tasks, paving the way for developing inclusive solutions that can also cater to minorities in the demographics who are typically heavily under-represented in the training data.

\section{Methods}
The core of our method is a visual in-context learning model called SegGPT \cite{wang2023seggpt}. The input to the model is a visual prompt which is constructed by stitching together example input-output image pairs. The test input image with a blank output image is appended to the prompt. The task of the model is to complete the prompt, i.e., predict the blank output image for the given test input image (Figure \ref{fig:fig1}) without updating any model weights. 

\subsection{SegGPT Training}
The segmentation task in SegGPT is formulated as an image in-painting task \cite{wang2023images}: given an input image, the prediction task is to inpaint the desired but missing output image. This allows for a standard image input and image output interface for the model, because of which the model can be trained on a large corpus of vision data irrespective of the actual vision task. So long as the input and output of the vision task can be represented as images, the model can leverage the dataset associated with the task for training.

The training of SegGPT is based on MIM \cite{he2022masked}. During training, two images from the same segmentation task are stitched into a larger image, along with their corresponding masks. MIM is then applied to the pixels of output mask images. The large masking ratio forces the model to gain contextual and semantic understanding of the image to complete the in-painting task \cite{he2022masked,xie2022simmim}. This understanding allows the trained model to understand what areas to segment from the example input-output pairs in the prompt. 

The model in SegGPT employs a vanilla vision transformer (ViT) \cite{dosovitskiy2020image} as the encoder consisting of stacked transformer blocks. A three layer head comprising of a linear layer, a 3 x 3 convolution layer, and another linear layer is used to map the features of each patch to the original resolution. A simple smooth $l_1$ regression loss is used to train the network.  

\subsection{Prompt Selection}
The performance of visual in-context learning models depends strongly upon the choice of prompts \cite{zhang2023makes}. Inspired from the results in \cite{zhang2023makes}, we adopt a similarity-based method to retrieve the prompt from the training dataset. Specifically, for each test image, we compute its $k$ nearest neighbors from the training dataset based on a distance metric such as Frobenius norm and structural similarity index (SSIM). These neighbors are then used to construct the prompts for the given test image. 

\section{Experiments and Results}
The goals of our experiments were to investigate whether visual in-context learning can perform as well as traditional methods that rely on training on large datasets. To that effect, we compared the performance of a pretrained SegGPT against a CNN-based U-Net trained to segment eczema images. 

\subsection{Dataset description}
Our dataset consists of 528 high-resolution images collected from two primary sources: a) public Dermnet dataset of eczema images \cite{DermNet}, and b) an in-house consumer study where images were taken directly by the participants using their smartphone cameras. The resulting dataset has multiple sources of variations: eczema images from different body parts, varying skin tones, and varying levels of eczema along with varying illumination, background, etc. which increases the complexity of the segmentation task. 

The dataset was labelled using human annotators. The human annotators were provided written instructions from domain experts on how to create the masks. Several examples were shown to them before they started the task. Each image was annotated by one human annotator. The resulting masks were then inspected by multiple domain experts and were found to be satisfactory. 

All images were resized to 448 x 448 which is the input dimension for the pretrained SegGPT model. Further, the images were normalized using z-score normalization using the ImageNet dataset statistics. The dataset was partitioned into 428 training images and 100 evaluation images. The baseline U-Net model used all 428 images for its training. The SegGPT model used a handful of $k$ samples from the training set to construct prompts. 

\subsection{Baseline for comparison}
As baseline, we used CNN-based U-Net which is widely used for skin segmentation \cite{ronneberger2015u}. The architecture was a typical 5-stage U-Net comprising of serial contracting and expansive paths \cite{ronneberger2015u}. Similar to \cite{ronneberger2015u}, the contracting path consisted of repeated two 3x3 unpadded convolutions, followed by batch normalization, ReLU and maxpooling with window size 2x2. The number of channels were doubled after every stage. The expansive path followed the inverse operations of the contracting path- upsampling of feature maps followed by repeated up-convolutions that halve the number of channels, with ReLU non-linearity.

The network was trained on 428 training images using Adam optimizer with a small weight decay factor for additional regularization. Learning rate was set to $1e-4$. Batch size was set to 8. The network was trained for 50 epochs with cross-entropy loss function. 

To evaluate, we use the intersection over union score (mIoU) averaged over all test images. As the name suggest, mIoU measures the area of overlap between the predicted segmentation mask and the ground-truth mask, and hence is a more suitable metric to measure segmentation performance than pixel-wise accuracy. 

\subsection{SegGPT Network Architecture and Prompt Retrieval}
We used an off-the-shelf pretrained SegGPT model that employs a vanilla ViT-large as its encoder. The model was trained to optimize the smooth $l_1$ regression loss. The authors in \cite{wang2023seggpt} pretrained the model on a large collection of benchmark segmentation datasets: ADE20K, COCO panoptic, Cityscapes, COCO semantic, LIP person, PASCAL VOC, PACO, iSAID and loveDA aerial, CHASEDB, DRIVE, HRF and STARE retinal vessel. 

For prompt selection, we selected $k$ nearest neighbor of each test image from the training dataset, varying $k$ from 1 to 15 (Figure \ref{fig:fig1}). We employed the following distance metrics: i) the commonly used Frobenius norm to compute euclidean difference between two matrices; ii) SSIM, which is a better indicator of similarity between images by virtue of being a perception-based metric.
\subsection{Segmentation of Eczema}
\begin{figure*}[t!]
    \centering
    \includegraphics[width=\textwidth]{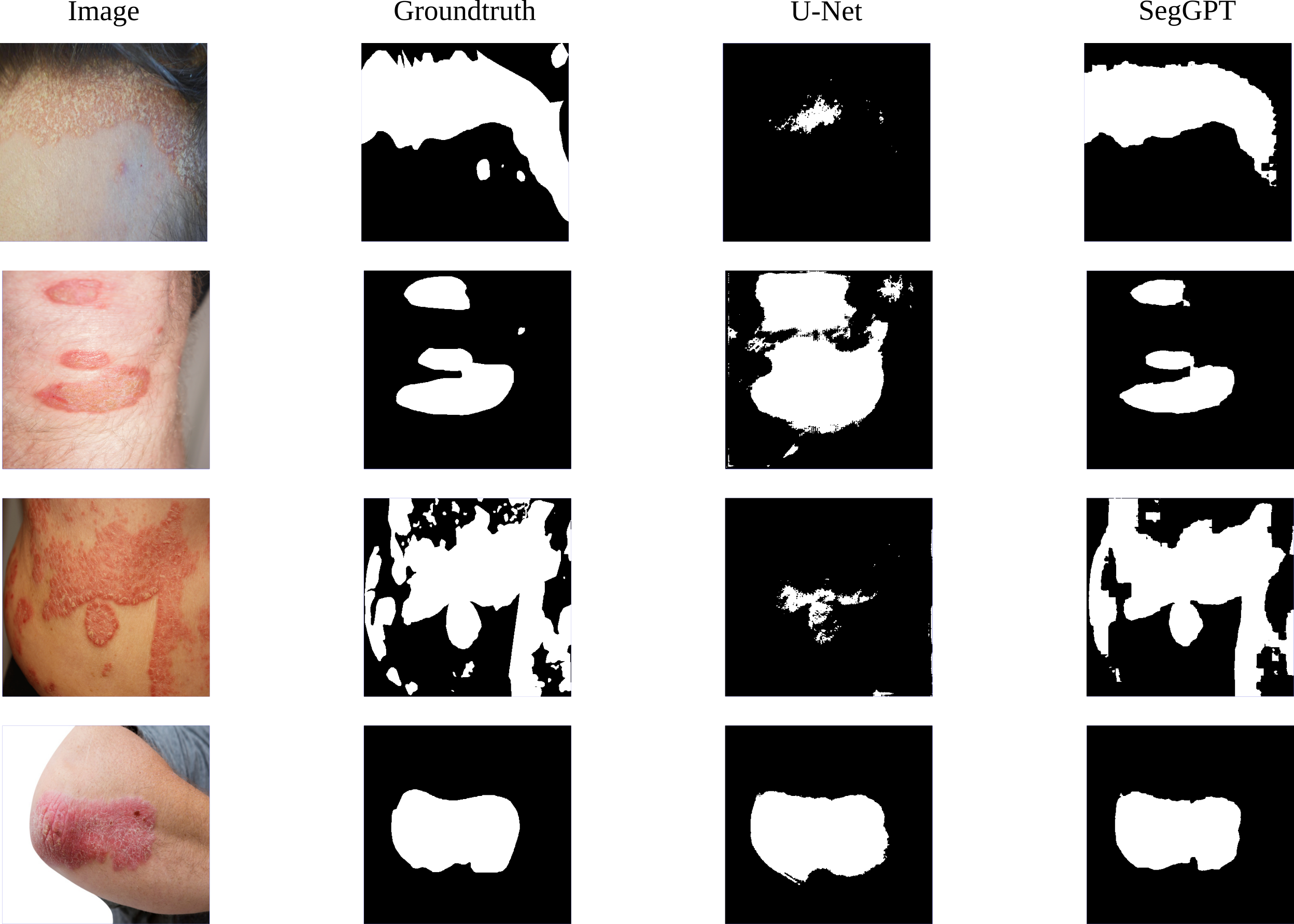}
    \caption{Qualitative comparison of eczema segmentation performance of U-Net and SegGPT for 4 example images. SegGPT produces masks that are closer to the groundtruth.}
    \label{fig:fig2}
\end{figure*}

\begin{table}
\small
  \caption{Segmentation Performance}
  \label{main-result}
  \centering
  \begin{tabular}{lc}
    \toprule
    Method & mIoU \\
    \midrule
    U-Net & 32.60\\
    SegGPT ($k=2$; SSIM) & 36.69  \\
    \bottomrule
  \end{tabular}
  \label{tab:performance}
\end{table}
We evaluated pretrained SegGPT and baseline U-Net for segmentation of eczema. While U-Net achieved an mIoU score of 32.60 on the test images, the pretrained SegGPT with the optimal hyperparameter $k=2$ achieved an mIoU score of 36.69 (Table \ref{tab:performance}). Not only did SegGPT outperform U-Net, it did so by using only 2 representative images from the training set and without changing any of its weights. This amounts to a 12.6\% increase in performance with a 213 times decrease in the training data requirement. The key is that the images in the prompt must be representative of the segmentation task. The quantitative improvement is further highlighted in the qualitative comparison in Figure \ref{fig:fig2}, where SegGPT produced comparable or higher quality masks than U-Net. 

\subsubsection{Effect of $k$ and distance metric}
\begin{figure*}
    \centering
    \includegraphics[width=0.55\textwidth]{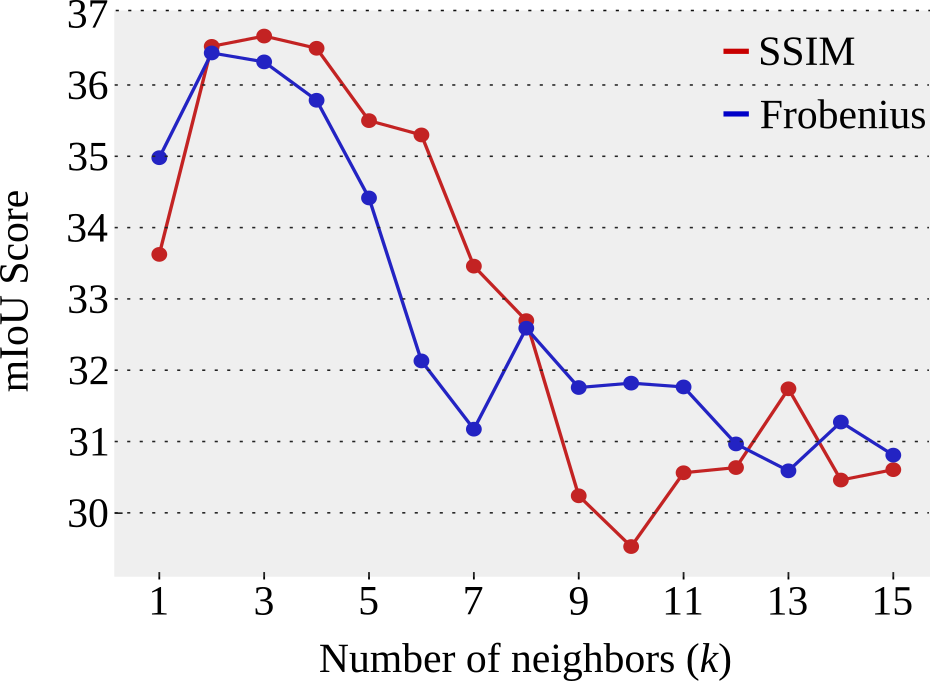}
    \caption{Dependence of the segmentation performance on the number of neighbors, $k$ and the distance metric. The performance increases up until a low value of $k$ and then decreases. Using SSIM as distance metric results in marginally better performance than using Frobenius norm.}
    \label{fig:fig3}
\end{figure*}

\begin{figure}
    \centering
    \includegraphics[width=\textwidth]{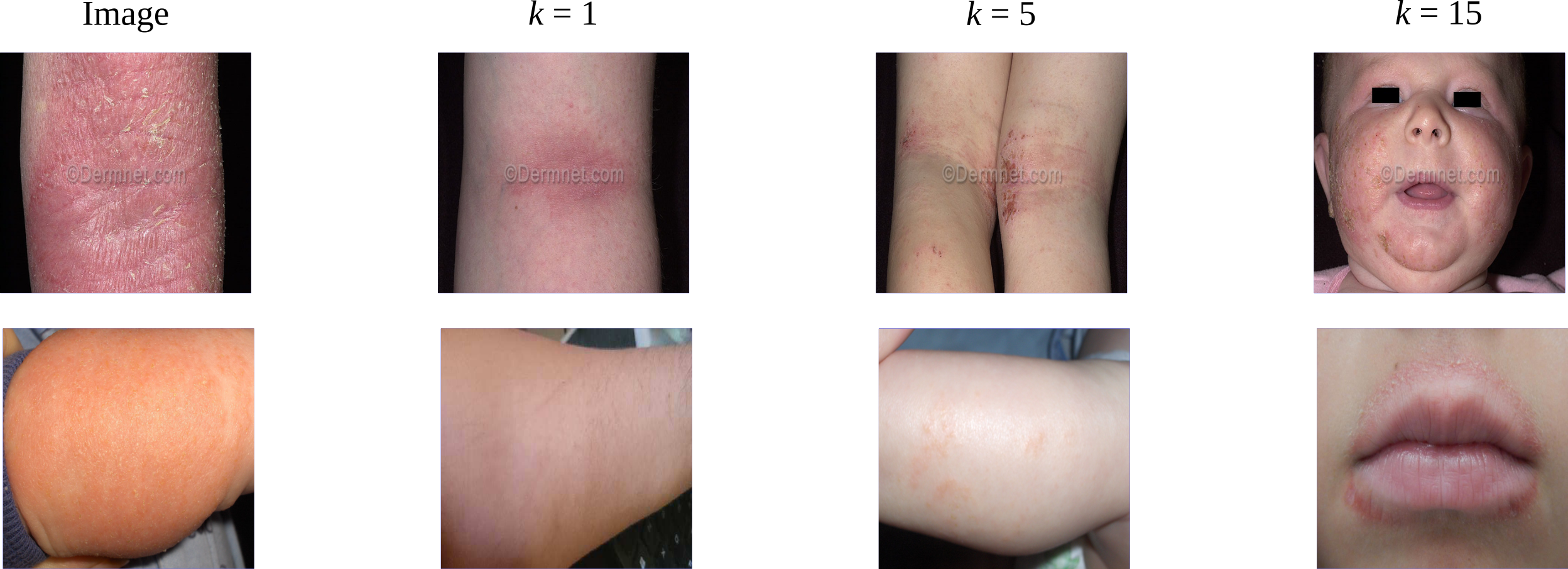}
    \caption{Nearest neighbor images for two example test images. As $k$ increases, the retrieved neighbor is further away from the test image.}
    \label{fig:fig4}
\end{figure}

To understand the dependence of the performance of SegGPT on the number of examples used to construct the prompt, we evaluated its performance for a range of values of $k$. We show in Figure \ref{fig:fig3} that the mIoU increases as we increase $k$ up to a certain level, and then starts to decrease. This indicates that picking higher number of examples to construct the prompt may not result in better performance. Although this may seem counter-intuitive, we believe that the reason for this observation is as follows: As we increase $k$, we pick examples that are further away from the test image. Given the limited size of the dataset, images that are too far from the test image may not be representative. To highlight this point, we show the $k$ nearest neighbors for two representative test images for both the distance metric in Figure \ref{fig:fig4}. Since the SegGPT model is not retrained/conditioned on the prompt, it is not capable of distinguishing examples that may not be representative of the segmentation task, and instead places equal emphasis on all the examples in the prompt. As one might expect, the performance obtained when SSIM is used as a distance metric is slightly better than when Frobenius norm is used (Figure \ref{fig:fig3}).

\section{Discussion}
In this work, we present a visual in-context learning approach for segmentation of eczema from patient-acquired images. We showed that with just two examples of the segmentation task in the prompt, and the right prompt-selection strategy, the pretrained SegGPT can perform better than the state-of-the-art CNN-based U-Net despite the fact that the latter sees 428 examples in its training. Our result adds to the mounting evidence that learning task-agnostic features on large diverse datasets with high-capacity models eliminates the need for performing any training or finetuning for downstream tasks \cite{liu2023summary,radford2021learning,oquab2023dinov2,kirillov2023segment}. 

A diverse dataset for medical and skin imaging that is representative of the underlying demographics is hard to get and even harder to annotate \cite{shi2019active}. The current methods to deal with the limited data use either self-supervision which relies on having enough unlabelled data, or finetuning which again requires sufficient labelled data \cite{tan2018survey}. On the other hand, visual in-context learning can leverage a handful of representative labeled examples to perform the task with competitive performance, drastically reducing the time and effort for data collection. While our current method uses the entire training dataset to search for representative examples, in a consumer-facing application, patients will need to provide just 1 or 2 annotated images for the model to make accurate predictions. 

In addition, the approach holds significance for application areas where it is impractical to wait for enough labelled data to arrive before accurate predictions can be made. Such application areas include consumer-facing applications where patients can self-monitor the trajectory of the improvement of their skin condition when following a treatment protocol \cite{horne1989preliminary,ehlers1995treatment,wittkowski2007beneficial}. In such cases, asking patients to wait until we have enough data from them may not be prudent, and in-context learning can play a key role.

Given the fact that the performance of in-context learning depends strongly on the choice of prompts, more systematic ways of prompt retrieval can be investigated. The current approach relies on pixel-level distance between the two images. However, measuring similarity between images at feature-level may result in more effective prompts. The work in \cite{zhang2023makes} presents two such approaches relying on supervised and unsupervised learning. Additionally, although the focus of this work is to obtain competitive performance without any training or finetuning, if additional performance improvement is desired, the ViT can further be finetuned on domain-specific data. 

A key issue in skin imaging is the under-representation of certain demographics in the training data, as a result of which methods may be biased towards heavily represented groups. In-context learning, with its ability to generalize from just a few data points, has the potential to tackle this issue. More experiments are needed on benchmark evaluation datasets that contain data from under-represented groups to confirm the hypothesis. Overall, our work highlights the importance of the increasingly-popular in-context learning framework, and the possible directional shift from the traditional train-test-finetune paradigm. 


%
%
%
\bibliographystyle{splncs04}
\bibliography{bib}

\end{document}